\documentclass[times, 10pt,twocolumn]{article}
\usepackage{latex8}
\usepackage{times}
\usepackage{amssymb}
\usepackage{amstext}
\usepackage{amsmath}
\usepackage{color}
\usepackage{eucal}
\usepackage{graphics}
\usepackage{graphicx}
\usepackage[T1]{fontenc}
\usepackage{ae,aecompl}

\begin{document}

\title{Automating the Dispute Resolution in a Task Dependency Network}
\author{Ioan Alfred Letia\\
Technical University\\
Department of Computer Science\\
Baritiu 28, RO-400391 Cluj-Napoca, Romania\\
letia@cs.utcluj.ro\\
\and
Adrian Groza\\
Technical University\\
Department of Computer Science\\
Baritiu 28, RO-400391 Cluj-Napoca, Romania\\
adrian@cs-gw.utcluj.ro\\
}

\maketitle
\thispagestyle{empty}

\begin{abstract}
When perturbation or unexpected events do occur, agents need
protocols for repairing or reforming the supply chain.
Unfortunate contingency could increase too much the cost of performance,
while breaching the current contract may be more efficient.
In our framework the principles of contract law are applied
to set penalties: expectation damages, opportunity cost, reliance damages,
and party design remedies, and they are
introduced in the task dependency model~\cite{Walsh03}.
\end{abstract}


\section{Introduction}
\label{sec:introduction}

The formation of the supply chain is grounded to three key technologies:
a) the decision-making mechanism of an individual business entity;
b) a coordination mechanism for the allocation of contracts;
c) the representation of capabilities and services.
In this paper we focus on the second aspect: contract allocation
Flexibility and risk sharing in supply contracting is a main issue
when confronted to the uncertain nature of the environment~\cite{Delft01}.
Contracts aim at improving flexibility,
while preserving the interests of both parties.
Our interest rely on remedies for breach of contracts.
An economic analysis of remedies models their effects on behavior.
The remedies imposed by low affect the agents' behavior~\cite{Cooter04}:
(1) searching for trading partners;
(2) negotiating exchanges;
(3) drafting the stipulations in contracts;
(4) keeping or breaking promises;
(5) taking precaution against breach causing events;
(6) acting based on reliance on promises;
(7) acting to mitigate damages caused by broken promises;
(8) settle disputes  caused by broken promises.
In the supply chain context a contract breach can propagate over the entire chain.
The damages imposed by legal institutions can positively influence breach propagation.
Usually, a contract breach appears when some perturbation  arises on the market\footnote{For instance, the market price of a raw material could rise so much that, for the agent who had planed to achieve it in order to produce an item, is more efficient to breach the contract with its buyer.}.
Therefore, proper rules imposed by the law help agents to manage perturbations in supply chain.
Each agent has more than one way to respond to a perturbation.
The question is how can we change the strategic environment such that the resulting behavior of the involved agents is efficient?
More exactly, what types of contracts or market structure can impose
this mutually acceptable solution if such a solution exists and,
in the same time, these contracts do not have to force the agents
to act irrationally when a mutual solution does not exist.
Our goal is to find which of the following remedies are adequate for an efficient functionality of the supply chain:
expectation damages $D_e$, opportunity cost $D_o$, reliance damages $D_r$, and party designed damages $D_p$.

The main contribution of this paper consists in introducing penalties in task dependency network model.
It also investigates the agents' decision to breach or perform the contract when they face different levels of information sharing.

In the next section we introduce contracts within the task dependency network model and in
section~\ref{sec:remedies} we describe the four types of remedies used in the market.
The section~\ref{sec:decision} discusses how breach decision is influenced
by information sharing between firms and
in section~\ref{sec:analyses} the functions used by the market for penalties are implemented.
Sections~\ref{sec:experiments} and~\ref{sec:related} detail future experiments and related work.

\section{Problem specification}
\label{sec:design}

\subsection{Task Dependency Network}

We adapted the task dependency network model~\cite{Walsh03} used in the
analysis of the supply chain as follows:
task dependency network is a directed, acyclic graph,
(V,E), with vertices $V = G \cup A$, where:
$G$ = the set of goods,
$A = S \cup P  \cup C$  the set of agents,
$S$ = the set of suppliers,
$P$ = the set of producers,
$C$ = the set of consumers,
and a set of edges E connecting agents with their input and output goods.
\begin{figure}
    \begin{center}
        \includegraphics[width=0.45\textwidth]{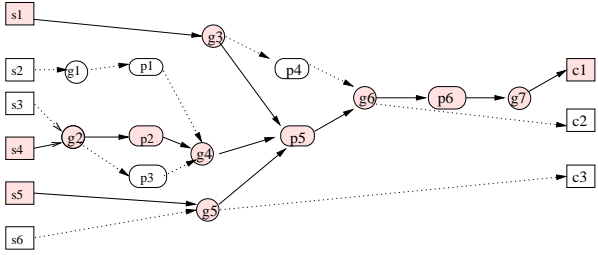}
    \end{center}
    \caption{Task dependency network: goods are indicated by circles,
suppliers and consumers are represented by boxes, while producers by curved boxes.}
    \label{fig:task}
\end{figure}
With each agent $a \in A$ we associate an input set $I_a$ and an output set $O_a$:
$I_a=\{g \in G |\prec g,a \succ \in E\}$ and $O_a=\{g \in G |\prec a,g \succ \in E\}$.
Agent $a$ is a supplier if $I_a=\emptyset$, a consumer if $O_a=\emptyset$, and a producer in all other cases. Without any generalization lost, we consider that a consumer $c \in C$ needs a single item ($|I_c|=1$) and every supplier $s \in S$ or producer $p \in P$ build one single item ($|O_s|=1$ and $|O_p|=1$)

An agent must have a contract for all of its input goods in order to produce its output,
named \emph{presumable}\footnote{Note that when someone breaches a contract with a presumable agent it has to pay more damages.} and denoted by $\hat{p}$.
If we note $n_p=|I_p|$, the agent has to sign $n_p+1$\footnote{Suppliers and consumers have to sign one contract only.} contracts in order to be a member in the supply chain.
 For each input good $g_k \in I_p$ the agent $p$ bids its own item valuation $v_p^k$.
The auction for the good $g_k$ sets the transaction price at $p_k$.
The agent's investments are $I_p=\sum_{k=1}^{n_p}{p_k}$ where $k$ are the winning input goods.
We note by $I_p^g$ the agent's investments but without considering the investments made for the current good $g$.
Similarly, we note all bids values submitted by the agent $p$ as $V_p=\sum_{k=1}^{n_p}{v_p^k}$ and this value without considering the bid for good $g$ as $V_p^g$.
For the output good, the agent $p$ signs a contract at reliance price $R_p$.

We consider that there are no production costs and
when perturbation or unexpected events occur,
agents need  protocols for repairing or reforming the supply chain.
"Allowing decommitment without remedies rises the question
of how to enforce that agents decommit only when they are in dead ends,
 and also does not address the fact that unilateral decisions for decommitment
can potentially break the (possibly desirable) contracts
of many other downstream producers"~\cite{Walsh03}.
Introducing remedies can reduce aggressive bidding and mitigate the potential problems.

\subsection{Contracts}

The goods are transacted using the (M+1)st price auction
protocol~\cite{Wurman98}, which has the property to balance the
offer and the demand at each level in the supply chain (otherwise
the supply demand equilibrium cannot be achieved globally). It
provides a uniform price mechanism: all contracts determined by a
particular clearing are signed at the same price. In the contract
$$C=\prec a_s, a_b, g_i, P_c, t_{issue}, t_{maturity}   \succ$$
$a_s$ represents the seller agent, $a_b$ the buyer agent, $g_i$ the good or the transaction subject, $P_c$ the contract price, $t_{issue}$ is the time when the contract is signed and $t_{maturity}$ is the time when the transaction occurs.
During experiments, a contract can be in one of the following states: active (between $t_{issue}$ and $t_{maturity}$ and no breach), violated (at the time of breach
$t_{issue} \leq t_{breach} \leq t_{maturity}$) or performed (if no party breaches until $t_{maturity}$).


\section{Remedies}
\label{sec:remedies}

According to~\cite{Vold02}, there are five different philosophies
of punishment from which all punishment policies can be derived:
deterrence, retribution, incapacitation, rehabilitation and
restoration. Retribution is most adequate for multi-agent
systems~\cite{Chaib-draa04}, as it considers that the contract
breach should be repaired by a remedy as severe as the wrongful
act\footnote{Courts call a term "liquidated damages" when it
stipulates damages that do not exceed the actual harm. When the
stipulates damages exceed the actual harm caused by breach, the
remedies are called penalties.}. The remedies described in this
section try to equal the victim's harm. In the first three
cases\footnote{Expectation damages, reliance damages and
opportunity cost are analyzed from an economical point of view
in~\cite{Cooter04,Friedman00}.}, the system estimates the harm
according to current market conditions, while in the last case,
the agents themselves compute the damages and generate their own
penalties.

\subsection{Expectation Damages}

The courts reward damages that place the victim of breach in the position he or she would have been in if the other party had performed the contract~\cite{Cooter04}.
Therefore, in an ideal situation, the expectation damages does not affect the potential victims whether the contract is performed or breached.
Ideal expectation damages remain constant when the promisee relies on the performance of the contract more than it is optimal.

\subsection{Reliance Damages}

Reliance increases the loss resulting from the breach of the contract.
Reliance damages
put the victim in the same position after the breach as if he had not signed a contract with the promisor or anyone else~\cite{Cooter04}.
In an ideal situation, the reliance damages do not affect the potential victims whether the contract is breached or there was no initial contract.
No contract provides a baseline for computing the injury.
Using this baseline, the courts may reward damages that place victims of breach in the position that they would have been, if they had never contracted with another agent.
Reliance damages represent the difference between profit if there is no contract and the current profit.

\subsection{Opportunity Cost}

Signing a contract often entails the loss of an opportunity to make an alternative.
The lost opportunity provides a baseline for computing the damage.
Using this baseline, the courts reward damages that place victims of breach in the position that they would have been if they had signed the contract that would have been the best alternative to the one that was breached~\cite{Cooter04}.
In the ideal situation, the opportunity cost damages does not affect the potential victims whether the contract is breached or the best alternative contract is performed\footnote{Opportunity cost and expectation damages approach equality as markets approach perfect competition.}.
If breach causes the injured party to purchase a substitute item, the opportunity cost formula equals the difference between the best alternative contract price available at the time of contracting and the price of the substitute item obtained after the breach.

\subsection{Party-Designed Remedies}

The contract might stipulate a sum of money that the breaker will
pay to the party without guilt. These "leveled commitment
contracts"~\cite{Sandholm01} allow self-interested agents to face
the events that unfolded since the contract started. A rational
person damages others whenever the benefit is large enough to pay
an ideal compensation and have some profit, as required to
increase efficiency. Game theoretic analysis has shown that
leveled committed contracts increase the Pareto efficiency of
contracts. One contract may charge a high price and offer to pay
high damages if the seller fails to deliver the goods, while
another contract may charge a low price and offer to pay low
damages, the types of contracts separating the set of buyers and
allowing "price discrimination."


\section{Efficient breach}
\label{sec:decision}

Breaching is more efficient than performing when the costs of performing exceed the benefits to all parties.
The costs of performing exceed the benefits when a contingency materializes so that it makes the needed resources for performance more valuable in an alternative usage.
Two types of contingencies reorder the value of resources: unfortunate contingency increases the cost of performance (an unpredictable strike) or fortunate contingency makes nonperformance even more profitable than performance
(the seller discovers a buyer who values the product even more).

The literature distinguishes between $a$ $priori$ and $a$ $posteriori$ decided sanctions.
According to~\cite{Chaib-draa04}, we consider that a posteriori decided penalties should be avoided in agent-based modeling, since they do not allow agents to reason about when to respect their commitments.
Therefore, all the above remedies $D_e$, $D_r$, $D_o$, and $D_p$ are common knowledge and settled a priori.
Nevertheless, only $D_p$ penalty fully reveals the amount of money the agents will pay in case of breach.
In the other cases, more information is needed for an agent in order to anticipate how high remedies he or she has to pay.
Hence, the potential victim could advertise how much he or she will lose depending on the types of remedies imposed by the market (see table~\ref{tab:InformationSharing}).
\begin{table}
    \begin{center}
                    \begin{tabular}{l|c}
                    Remedy & Information shared\\
                    \hline
                    $D_e$ & expected profit\\
                    \hline
                    $D_o$ & provided by the market\\
                    \hline
                    $D_r$ & investments made\\
                    \hline
                    $D_p$ & known from contract\\
                    \hline
                    \end{tabular}

    \end{center}

    \caption{Dependence of shared information
    }
    \label{tab:InformationSharing}
\end{table}
This approach stresses the relation between legal rules, communication and common knowledge. 
We define three levels of information sharing as in [17] (no share, share to each neighbor, and broadcast sharing) and we will analyze the social welfare in each case.
The market should provide incentives for information sharing.
In our case, hiding such information in order to collect more penalties should not be encourage by the remedies rules.


\section{Case Analysis}
\label{sec:analyses}

The conclusions from the last sections are:
(i) The amount of expectation damages must place the victim
in the same position as if the actual contract had been performed
\footnote{We assume that the rate of breach is low.
Otherwise, it can be anticipated to some extent,
and so the promisee can plan for breach, just as airlines and
hotels plan for "no-shows~\cite{Cooter04}."}
(ii) The amount of reliance damages must place the victim in the
same position as if no contract had been signed; (iii) The amount
of opportunity-cost damages must place the victim in the same
position as if the best alternative contract had been performed;
(iv) Party designed remedies specify themselves the amount of
damages in case of a breach.

\subsection{No substitute}

\subsubsection{Supplier-Consumer:}

\begin{figure}
    \begin{center}
        \includegraphics[width=0.45\textwidth]{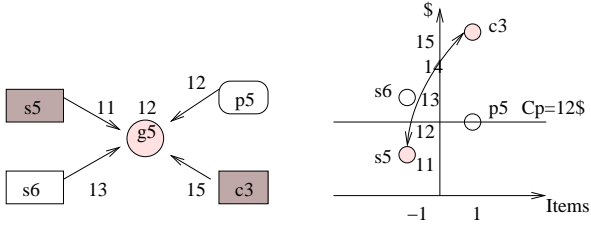}
    \end{center}
    \caption{Supplier-Consumer contract}
    \label{fig:supplier-consumer}
\end{figure}
First, we analyze the easiest case in which a contract is signed
between a supplier and a client. Recall that a supplier is an
agent who does not have any input good in task dependency network
model, while the consumer is an agent who does not have any output
good.

\paragraph{\textbf{The consumer breaches the contract:}}
In fig.~\ref{fig:supplier-consumer}a) the suppliers $s_5$ and
$s_6$ want to sell good $g_5$ at price 11 and respectively 13,
while the agents $p_5$, and $c_3$ try to buy it at prices 12 and
15. According to (M+1)st price protocol the transaction price is
$P_c=12\$$. The auction clears at every round. In
fig.~\ref{fig:supplier-consumer}b) a single contract is signed:
$C_{g_5}^1=\prec s_5,c_3,g_5,12,t_{issue},t_{maturity}\succ$
Consider that $c_3$ breaches the contract. In this case, the
remedies will be:
\\\emph{Expectation damages}:
if the agent $c_3$ performs, the $s_5$'s estimated profit is the difference between the contract price $P_c=12$ and its own valuation\footnote{(M+1)st price auction has the following property: the dominant strategy for each agent is to reveal its real valuation.}
$v_{a_6}^{g_5}=11$ (victim valuation).
The remedies compensate this value:
$$D_e=P_c-v_a^g$$
\emph{Opportunity damages}: first, the auctioneer has to compute
the opportunity cost $P_o$, which is the transaction cost in case
the breacher was absent from the auction. In
fig.~\ref{fig:supplier-consumer}, if agent $c_3$ is not present
$P_o=11$. The $s_5$'s bid is one who wins. The contract would be
$C_{g_5}^1=\prec s_5,p_5,g_5,11,t_{issue},t_{maturity}\succ$ and
the agent's profit would be $P_o-v_a^g$. But, when there is no
contract for agent $s_5$, his profit would be null. The
opportunity damages should reflect this. We define opportunity
cost damage $D_o$ which is received by the agent $a$ as:
$$D_o=max(P_o-v_a^g,0)$$
\emph{Reliance damages}: if the victim does not have any input
good, the supplier's investments in performing are null: $D_r=0$.
\\\emph{Party-designed remedies}: the remedies may be a fraction from the contract price ($D_p=\alpha\cdot P_c$) or a fraction from the expected profit ($D_p=\alpha\cdot D_e$) or constant ($D_p=C)$.
In each of the following cases, this type of remedies is computed in the same manner.
{\paragraph{\textbf{The supplier breaches the contract:}}

Consider that $s_5$ breaches contract $C_{g_5}^1$.
\\\emph{Expectation damages}:
$D_e=v_a^g-P_c$.
\\\emph{Opportunity damages}: if the breacher had not bid and the victim had signed a contract at the opportunity price $P_o$, than it's profit would have been $v_a^g-P_o$.
If the victim has no contract when the breacher is not bidding, it receives no damages.
Hence
$$D_o=max(v_a^g-P_o,0)$$
In the depicted case, if the agent $s_5$ had not existed, $c3$ would have signed a contract with $s_6$ for an opportunity cost $P_o=12$. Therefore, $D_o=3$.
\\\emph{Reliance damages}: because the client does not produce any output goods,
it's reliance is null: $D_r=0$.
\subsubsection{Supplier-Producer}
\label{subsec:supplier-producer}
\begin{figure}
    \begin{center}
        \includegraphics[width=0.45\textwidth]{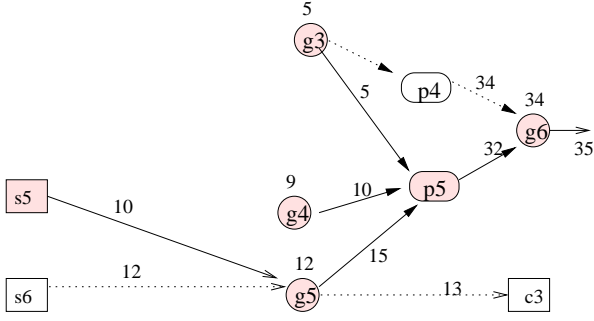}
    \end{center}
    \caption{Supplier-Producer contract}
    \label{fig:supplier_producer}
\end{figure}
\paragraph{\textbf{The supplier breaches the contract:}}
Consider $C=\prec s_5, p_5,g_5,12,t_{issue},t_{maturity}\succ$ from fig.~\ref{fig:supplier_producer}.
\\\emph{Expectation damages:}
Observe that the victim is a presumable agent because it has contracts for all its input goods.
Its investments are $I_p=5+9+12=26$ and $I_p^{g_5}=9+5=14$.
The producer $p_5$ has also a contract for its output item, so $R_{p_5}=34$.
Its profit is $R_{p_5}-I_p=8$.
When bad contracts have been signed this value can be negative, therefore no damages are imposed.
Otherwise, expectation damages equals the difference between its bid and the contract price:
$$D_e=
\begin{cases}
max(R_p-I_p,0), & \text{$\hat{p}$, $\exists{R_p}$}\\
v_p^g-P_c,  &\text{otherwise}
\end{cases}
$$
Recall $\hat{p}$ means that agent $p$ is presumable.
\\\emph{Opportunity cost}: one seller less implies $P_o\geq P_c$.
$$D_o=
\begin{cases}
max(R_p-I_p^g-P_o,0),   & \text{$\hat{p}$, $\exists{R_p}$, $\exists{P_o}$}\\
v_p^g-P_o,  &\text{$\neg \hat{p}$, $\exists{R_p}$, $\exists{P_o}$}\\
0,  &\text{$\neg{\exists{P_o}}$}
\end{cases}
$$
which is equivalent to:
$$D_o=
\begin{cases}
max(R_p-I_p^g-P_o,0),   & \text{$\hat{p}$,$\exists{R_p}$, $\exists{P_o}$}\\
max(v_p^g-P_o,0),   &\text{otherwise}
\end{cases}
$$
\emph{Reliance damages}:
$$D_r=
\begin{cases}
V_p^g-I_p^g+R_p-v_p^{g_o},  & \text{$\hat{p}$, $\exists{R_p}$}\\
V_p^g-I_p^{g_k},    &\text{otherwise}
\end{cases}
$$
Here $g_o$ is the output good of the agent $p$ and $I_p^{g_k}$
represents all $k$ contracts signed for input goods, where
$k<n_p$. In the depicted case $p_5$ is presumable and there is a
contract with a buyer. Therefore, it has to receive, as a victim,
the next reliance damages
$D_r=V_{p_5}^{g_5}-I_{p_5}^{g_5}+R_{p_5}-v_{p_5}^{g_6}=(10+5)-(9+5)+34-32=3$.

In some cases damages can be higher than the contract value itself
($D_r \geq P_c$). According to current practice in law, these
damages are the right ones if the victim gives a previous
notification about the risks faced by the potential breacher. This
a clear situation when information propagation improves the supply
chain performance. In the light of the above facts, their reliance
damages should remain the mentioned ones if the victim has
notified its partner, but should be maxim $P_c$ otherwise. Hence,
we define $D'_r$:
$$D'_r=
\begin{cases}
D_r,    & \text{the breacher receives a notice}\\
min(D_r,P_c),   &\text{otherwise}
\end{cases}
$$

\subsubsection{Producer-Consumer}
\begin{figure}
    \begin{center}
        \includegraphics[width=0.45\textwidth]{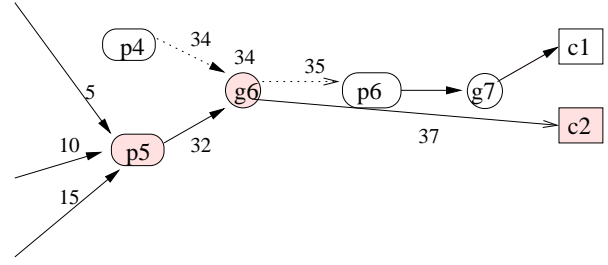}
    \end{center}
    \caption{Producer-Consumer contract}
    \label{fig:producer_consumer}
\end{figure}

\paragraph{\textbf{The consumer breaches the contract:}}
Consider the contract $C=\prec p_5, c_2,g_6,34,t_{issue},t_{maturity}\succ$ from fig.~\ref{fig:producer_consumer}, where $c_2$ breaches.
\\\emph{Expectation damages}:
$$D_e=
\begin{cases}
max(P_c-I_p^g,0),   & \text{$\hat{p}$}\\
P_c-v_p^g,  &\text{otherwise}
\end{cases}
$$
In this case, $p_5$ is presumable and $D_e=34-(12+9+5)=8$.
Suppose the agent $p_5$ does not have any contract for one input good.
Therefore, it is not presumable and it will receive $D_e=34-32$.
\\\emph{Opportunity cost}: one buyer less implies $P_o\leq P_c$
$$D_o=
\begin{cases}
max(P_o-I_p^g,0),   & \text{$\hat{p}$, $\exists{P_o}$}\\
P_o-v_p^g,  &\text{$\neg \hat{p}$, $\exists{P_o}$}\\
0,  &\text{$\exists \overline{P_o}$)}
\end{cases}
$$
which is equivalent to:
$$D_o=
\begin{cases}
max(P_o-I_p^g,0),   & \text{$\hat{p}$, $\exists{P_o}$}\\
max(P_o-v_p^g,0),   &\text{otherwise}
\end{cases}
$$
\emph{Reliance damages}:
$D_r=V_p-I_p$.

\subsection{Substitute}

The common law requires the promisee to mitigate damages.
Specifically, the promisee must take reasonable actions to reduce losses occurred by the promisor's breach.
The market can force the victim to find substitute items, in this case the imposed damages reflect only the difference between original contract and substitute contract.
With a substitute contract, the victim signs for the identical item, with the same deadline or $t_{maturity}$, but at a different price.
Let $P_s$ be the value of the substitute contract\footnote{$P_s$ comes from "spot market" while the original contract value refers to "future market".}.
For the general case \emph{Producer-Producer}, when \textbf{the buyer breaches the contract}, the equations become:
\\\emph{Expectation damages}:
$$D_e=
\begin{cases}
max(P_c-I_p^g,0),   & \text{$\hat{p}$, $\neg  \exists P_s$ }\\
P_c-v_p^g,  &\text{$\hat{p}$, $\neg  \exists P_s$}\\
max(P_c-P_s), &\text{$\exists P_s$}
\end{cases}
$$
\emph{Opportunity cost}:
$$D_o=
\begin{cases}
max(P_o-I_p^g,0),   & \text{$\hat{p}$, $\exists{P_o}$, $\neg  \exists P_s$}\\
max(P_o-v_p^g,0),   &\text{$\neg \hat{p}$,$\exists{P_o}$, $\neg  \exists P_s$ }\\
max(P_o-P_s,0), &\text{$\exists P_s$}
\end{cases}
$$
\emph{Reliance damages}:
$$D_o=
\begin{cases}
V_p-I_p,    & \text{$\neg  \exists P_s$}\\
max(P_c-P_s,0), &\text{$\exists P_s$}
\end{cases}
$$

\section{Planned experiments}
\label{sec:experiments}

First, the framework can be used as a tool for automated online
dispute resolution (ODR). There are three situations:
(i) The market may have substantial authority, hence \emph{one
remedy is imposed to all agents}. In this case, the amount of
penalties can be automatically computed with this framework.
(ii)
Consistent with party autonomy, \emph{the agents may settle on
different remedies at contracting time}. This approach increases
the flexibility  and efficiency, because the agents are the ones
who know what type of remedy protects better their interests.
(iii) \emph{All the above remedies influence the amount of
penalties}: in this approach the role of the framework is to monitor
the market and collect data such as: the expected profit, the
opportunity cost, the amount of investments made, if there is a
substitute at $t_{breach}$. All these information are used as
arguments when the dispute is arbitrated~\cite{Toulmin58} in an architecture which combine rule base reasoning (laws) and case base reasoning (training set) as fig~\ref{fig:argumentation}.

Second, knowing the bids, the actual contracts, the amount of
potential remedies, and the available offers on the market, the
framework can identify situations in which for both agents is more
profitable to breach the contract when a fortunate or an
unfortunate contingency appears. It computes pairs of suggestions,
helping to increase total welfare towards Pareto frontier.

Third, as a simulation tool, the market designer may obtain
results regarding the following questions: what types of remedies
assured flexibility in the supply chain? or
how information sharing influences total revenues or can be use to
compute optimum reliance? In the prototype developed we are
currently making experiments with different types of agents:
low-high reliance, breach often-seldom, sharing information-don't
share, risk seeking-averse (when they are risk averse, the
penalties do not need to be so high to make breachers behave
appropriately).
\begin{figure}
    \begin{center}
        \includegraphics[width=0.45\textwidth]{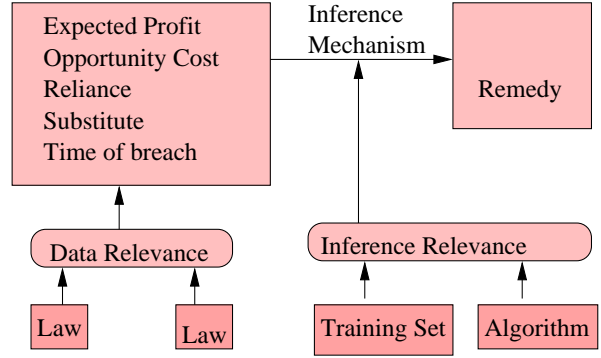}
    \end{center}
    \caption{Toulmin argument structure}
    \label{fig:argumentation}
\end{figure}


\section{Related Work}
\label{sec:related}
The task dependency network model was
proposed~\cite{Walsh03} as an efficient market mechanism in
achieving supply chain coordination. However, this approach is
rather a timeless-riskless economy. On real markets a firm seldom
signs contracts with its buyers and its suppliers simultaneously.
Moreover, the breach of a contract implies no penalties, which is
an unrealistic assumption in real world. In contrast, in our model
we used auctions, which end independently, and we introduce
penalties in case of contract breaching. We have made less
strictly the assumption that the producer places its first output
offer only after receiving the first price quotes for all its
inputs. One objective is to compute how high these penalties
should be in order to maintain efficiency of the supply chain.

Whether selfish companies have individual incentive to use
information sharing for reducing the bull-whip
effect~\cite{Moyaux04} has been studied
considering three levels of information sharing: not share, share
to each neighbor, and broadcast sharing. Instead, we have used the
same levels of information sharing in order to help agents decide
when to breach. 
The agents in our framework sign simple formal contracts, but more complex contracts need semantic interpretation for automated ODR~\cite{Grosof2}. 

The role of sanctions in multi-agent systems~\cite{Chaib-draa04}
is the enforcement of a social control mechanism for the
satisfaction of commitments. We focus only on material sanctions
and we do not include social sanctions which affect trust,
credibility or reputation. Moreover, we have applied four types of
material remedies in a specific domain. In the same spirit of
computing penalties according to the level of harm produced, the
amount of remedies may depend on the time when the contract was
breached~\cite{Toledo01}. 
Expectation damages, reliance damages,
and opportunity have also been
studied~\cite{Craswell00,Cooter04,Friedman00} and how contracts influence the supply chain coordination or strategic breach appear in~\cite{Cachoon00,Sandholm01}.

The Toulmin argument structure was used in a framework for computing the distribution of matrimonial property~\cite{Bellucci03}. The domain was modelled by extracting the relevant variables with the help of experts and a neuronal network was used as inference mechanism. We apply principles of contract law to determine the amount of remedies and, in our business scenario, data used for argumentation can be automatically extracted from the task dependency network.  

\section{Conclusions}
\label{sec:conclusions}

The design of punishment policies applied to specific domains
linking agents actions to material penalties is an open research issue~\cite{Chaib-draa04}.
The contribution of this paper contains two ideas:
On the one hand, we apply the principles of contract law
in the task dependency network model~\cite{Walsh03}.
As a result, we enrich that model by including different types of penalties when agents breach,
thus bringing the model closer to the real world.

On the other hand, in our work we consider penalties that ensure a
higher welfare for the supply chains affected by perturbations,
with three levels of information sharing for each type of remedy.
This framework is useful for automated ODR. The data obtained can be used as arguments in a mediated dispute or the remedies can be computed in real time in case the agents agreed with the market policy. 

\bibliographystyle{latex8}
\bibliography{pp}

\end{document}